\documentclass[pdflatex,sn-basic,Numbered,iicol,twocolumn]{sn-jnl}% KI/Springer layout with numbered citations and two-column formatting

\usepackage{graphicx}%
\usepackage{amsmath,amssymb,amsfonts}%
\usepackage{amsthm}%
\usepackage{mathrsfs}%
\usepackage[title]{appendix}%
\usepackage{xcolor}%
\usepackage{textcomp}%
\usepackage{booktabs}%
\usepackage[inline]{enumitem}

\begin{document}

% \title[StretchBot: Neuro-Symbolic Framework for Adaptive Assistive Robot Guidance]{StretchBot: Neuro-Symbolic Framework for Adaptive Assistive Robot Guidance}
\title{StretchBot: A Neuro-Symbolic Framework for Adaptive Guidance with Assistive Robots}

\author[2]{\fnm{Luca} \sur{Vogelgesang\textsuperscript{*}}}\email{luca.vogelgesang@etu.umontpellier.fr}

\author[2]{\fnm{Mehdi} \sur{Soltani\textsuperscript{*}}}\email{soltaniahmedmehdi@gmail.com}
\equalcont{\textsuperscript{*}These authors contributed equally to this work. 

\textsuperscript{†}Principal Investigator / Senior supervising author.}

\author[1]{\fnm{Mohammadhossein} \sur{Khojasteh}}

\author[1]{\fnm{Xinrui} \sur{Zu}}

\author[1]{\fnm{Stefano} \sur{De Giorgis}}

\author[2]{\fnm{Madalina} \sur{Croitoru}}

\author[1]{\fnm{Filip} \sur{Ilievski\textsuperscript{$\dagger$}}}

\affil[1]{\orgdiv{Faculty of Science, Department of Computer Science}, \orgname{Vrije Universiteit Amsterdam}, \orgaddress{\street{De Boelelaan 1105}, \city{Amsterdam}, \postcode{1081 HV}, \country{Netherlands}}}

\affil[2]{\orgdiv{Faculty of Science, Department of Computer Science and Robotics}, \orgname{University of Montpellier}, \orgaddress{\street{30 Pl. Eugène Bataillon}, \city{Montpellier}, \postcode{34095}, \country{France}}}

\abstract{
Assistive robots have growing potential to support physical wellbeing in home and healthcare settings, for example, by guiding users through stretching or rehabilitation routines. However, existing systems remain largely scripted, which limits their ability to adapt to user state, environmental context, and interaction dynamics. In this work, we present StretchBot, a hybrid neuro-symbolic robotic coach for adaptive assistive guidance. The system combines multimodal perception with knowledge-graph-grounded large language model reasoning to support context-aware adjustments during short stretching sessions while maintaining a structured routine. To complement the system description, we report an exploratory pilot comparison between scripted and adaptive guidance with three participants. The pilot findings suggest that the adaptive condition improved perceived adaptability and contextual relevance, while scripted guidance remained competitive in smoothness and predictability. These results provide preliminary evidence that structured actionable knowledge can help ground language-model-based adaptation in embodied assistive interaction, while also highlighting the need for larger, longitudinal studies to evaluate robustness, generalizability, and long-term user experience.

}


\maketitle
\section{Introduction}
\label{sec:intro}

Robotic systems are increasingly deployed in domestic and healthcare environments to support human wellbeing through physical assistance, rehabilitation, and socially assistive interaction \cite{Broadbent2009,Tapus2007}. In rehabilitation and assistance contexts, such systems have shown potential to support motor functionality, training, and daily activities \cite{Mohebbi2020,Banyai2024}. However, many existing systems remain rigid and insufficiently adaptive to users' behavior, routines, and individual needs, offering limited personalization based on the user's physical state, emotional condition, or surrounding context \cite{Mohebbi2020}. This lack of adaptation can reduce engagement, hinder safety, and limit long-term adoption.

The fundamental challenge of creating an adaptive system lies in balancing two competing demands: (1) \textit{structured predictability}—a clear set of instructions that ensures comprehensive exercise coverage and that the user can confidently follow the routine, and (2) \textit{contextual responsiveness}—real-time adaptation to the user's current state, emotions, and environment. Purely scripted systems excel at predictability but fail to personalize and anticipate. Fully reactive systems can adapt but risk incoherence and repeated exercises.

To address these challenges, we attempt to answer three core research questions:

\noindent \textbf{RQ1:} How can we develop a hybrid neuro-symbolic architecture
% , connecting perception, reasoning with LLMs and knowledge graphs (KGs), and action execution 
to enable effective plan adaptation in assistive robotics?

\noindent \textbf{RQ2:} How should actionable knowledge be represented and integrated with LLM reasoning to ground robot decisions in both perception and domain knowledge?

\noindent \textbf{RQ3:} How do users perceive adaptive robot guidance in comparison to scripted routines along different dimensions?

To address these questions, we developed \textbf{StretchBot}, an adaptive robotic coach that guides users through a short stretching routine while dynamically adjusting its behavior in response to real-time multimodal cues. StretchBot integrates perception, commonsense reasoning, and interaction control in a unified architecture, enabling context-aware adaptation of stretching plans. At its core lies a \textit{neuro-symbolic reasoning module} that couples a domain knowledge graph (KG) with a large language model (LLM) for decision support, in line with commonsense reasoning methods \cite{ilievski2024human}. An additional verifier module acts as an optional safeguard stage to check response formatting and tone before execution in the current prototype.

To complement the technical system description, we report a small-scale pilot comparison between scripted and adaptive guidance. The goal of this preliminary study is not to provide statistically grounded validation, but to examine feasibility in real interactive sessions, collect early user impressions, and identify design trade-offs that motivate a larger controlled evaluation.

Answering our research questions, the main contributions of this work are:

\begin{enumerate}
\item \textbf{An adaptive stretching robot framework} combining multimodal perception with a hybrid neuro-symbolic planning architecture, alongside a collaborative interaction protocol that ensures explicit user confirmation to balance proactive adaptation with human oversight.

\item \textbf{A methodology for representing and integrating knowledge graph reasoning within LLM prompt design}, effectively grounding the robot's decisions in perceived user state, environmental constraints, and interaction history.

\item \textbf{An exploratory pilot evaluation} comparing scripted versus adaptive robotic guidance to identify early usability trade-offs in assistive stretching, complemented by an analysis of the strengths and limitations of hybrid LLM-KG planning.
\end{enumerate}
\section{Related Work}
\label{sec:related}

Our work sits at the intersection of four research areas: assistive robotics, commonsense reasoning, multimodal interaction, and LLM-based planning.

\paragraph{Assistive Robotics for Physical Guidance}

Robots have been explored as assistants for rehabilitation and exercise support, with the goal of improving user wellbeing and engagement \cite{Fasola2013,Mohebbi2020}. However, many prior systems rely on fixed scripts or predefined adaptation rules, which can limit responsiveness to the user's state, behavior, and surrounding context \cite{Mohebbi2020}. These limitations point to three recurring challenges in assistive robotics: ensuring safety while granting autonomy, personalizing guidance to individual needs, and maintaining user engagement over time. StretchBot addresses this gap by coupling a structured exercise baseline with adaptive branching points, allowing the system to adjust instructions during a session while preserving routine coherence.

\paragraph{Commonsense Reasoning and KGs}

Commonsense reasoning has been used to give robots a better understanding of their environment, and user needs \cite{Hogan2021}. Knowledge graphs provide structured, interpretable representations of entities and relations, and have become an important mechanism for grounding AI systems with explicit knowledge \cite{Hogan2021,ilievski2024human}. Public resources such as ConceptNet encode broad everyday relations that can support commonsense-aware reasoning \cite{Speer2017}. Recent work on retrieval-augmented generation has shown that combining external knowledge sources with language models can improve grounding and transparency in reasoning \cite{Lewis2021}. Building upon this idea for real-time adaptive guidance, StretchBot couples KGs with LLMs to leverage both the explainability of symbolic systems and the flexibility of neural language models. In our design, commonsense reasoning connects perceived user states—such as fatigue or thirst—with relevant object affordances and situation-specific suggestions, enabling the robot to decide proactively when to suggest a break or indicate a helpful resource (e.g., water, coffee).

\paragraph{Multimodal Human-Robot Interaction}

Prior work has shown that combining multiple sensing modalities, including visual, vocal, and contextual cues, can support emotion-related interpretation and improve the overall quality of human-robot interaction \cite{Tsiourti2019,Rawal2022}. For example, facial expressions and vocal signals can provide complementary information about the user's affective state in HRI settings \cite{Tsiourti2019,Rawal2022}. At the same time, body pose estimation can support real-time monitoring of posture and movement during exercise or rehabilitation tasks \cite{Bazarevsky2020}. These findings suggest that multimodal interaction can benefit from combining complementary sensing channels rather than relying on a single source of evidence. This motivates our use of optional affective channels alongside core perception components, such as posture monitoring and object detection, while keeping the final reasoning grounded in the most task-relevant signals.

\paragraph{LLM-Based Planning and Reasoning}

Recent work has explored the use of LLMs for robotic planning and decision-making \cite{Ahn2022, Brown2020}. The seminal SayCan work \cite{Ahn2022} combined LLM-based reasoning with affordance detection to generate feasible action plans. Subsequent research \cite{Lewis2021} has demonstrated that retrieval-augmented generation—augmenting LLMs with structured knowledge bases—improves both factual grounding and reasoning transparency. These insights reveal that while LLMs excel at flexible, context-aware reasoning, they require grounding mechanisms to keep their suggestions aligned with domain constraints. 
StretchBot combines LLMs with a domain-specific knowledge graph and an optional verification layer, but differs from prior LLM-for-robotics approaches by emphasizing an \textit{action-oriented knowledge representation} designed for runtime use rather than post-hoc context enrichment. This design grounds LLM outputs in structured knowledge and a constrained action format, making the resulting decisions easier to map to embodied robot actions in assistive settings.

\begin{figure*}[t]
\centering
\includegraphics[width=0.75\textwidth]{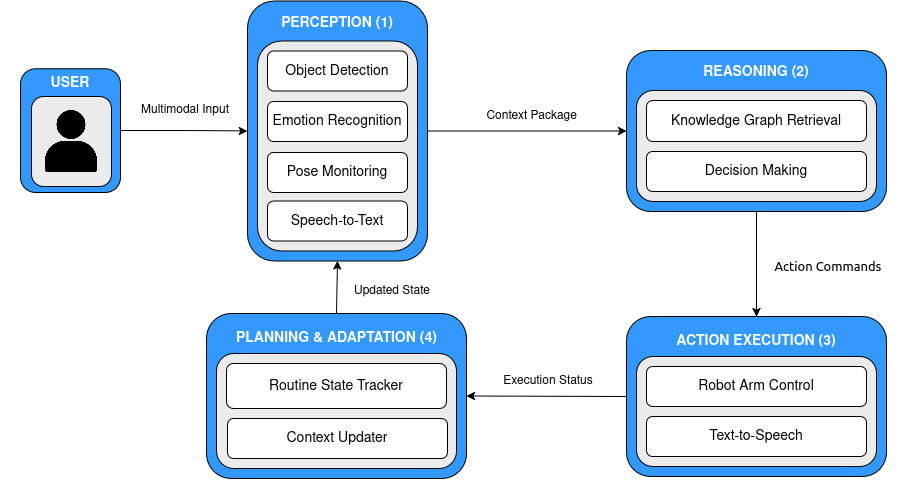}
\caption{The StretchBot pipeline is organized into four stages: Perception, Reasoning, Action Execution, and Planning and Adaptation. The arrows indicate the main information flow between blocks.}
\label{fig:overall_pipeline}
\end{figure*}

\section{System Architecture}
\label{sec:architecture}

% StretchBot follows a four-stage pipeline: Perception, Reasoning, Action Execution, and Planning \& Adaptation. 
Figure~\ref{fig:overall_pipeline} summarizes the 
% closed control loop and the direction of information flow of
% StretchBot, consisting of 
four stages of StretchBot: Perception, Reasoning, Action Execution, and Planning \& Adaptation.
Each stage transforms raw inputs into increasingly higher-level representations, forming a continuous evaluation-and-response loop that allows the robot to adjust its stretching instructions dynamically based on immediate multimodal feedback. 
% Figure~\ref{fig:overall_pipeline} summarizes the main information flow in StretchBot. 
The \textit{Perception} stage receives multimodal input from the user and extracts relevant features to assemble a context package, including detected objects, exercise status, optional affect cues, and dialogue input. This context is passed to the \textit{Reasoning} stage, where KG retrieval and LLM reasoning are combined to generate action commands such as \texttt{NEXT\_EXERCISE}, \texttt{POINT\_<OBJECT>}, or \texttt{STOP\_ROUTINE}. These commands are then executed by the \textit{Action Execution} block, and the resulting execution feedback is sent to the \textit{Planning \& Adaptation} block, which updates the current routine state and closes the adaptive loop.

\subsection{Perception}

% As illustrated by the \textit{Perception} block in Figure~\ref{fig:overall_pipeline}, 
The perception module corresponds to the first stage of the adaptive loop. It provides the adaptive system with a continuous and multimodal understanding of both the user's state and the environment. It integrates four complementary components: object detection, multimodal emotion recognition, pose monitoring, and speech-to-text.

\paragraph{Object Detection}
The object detection component continuously scans the robot's visual field to maintain an up-to-date representation of the physical environment. Its primary role is to identify and track objects that could be relevant to the user's well-being or the execution of the stretching routine (e.g., a chair for resting, a glass of water for hydration). By maintaining a dynamic inventory of the surroundings, this module allows the system to ground its reasoning in the actual environment, enabling the robot to make proactive, context-aware suggestions (e.g., pointing to a nearby water bottle when the user expresses fatigue) rather than relying on predefined assumptions.

\paragraph{Emotion Recognition}
Understanding the user's affective state is critical for tailoring the pace and tone of the interaction, as well as for proposing suitable helpful items, ensuring that the robotic coach remains supportive and responsive to signs of fatigue, frustration, or discomfort. To achieve this, the emotion recognition component captures and fuses affective signals across multiple modalities, providing a unified estimation of the user's current emotional state.
This component extracts emotions across three complementary channels:
\begin{enumerate*}
\item \textit{voice emotion recognition}, by analyzing the acoustic features and prosody of the user's speech;
\item \textit{facial emotion recognition}, by evaluating facial expressions and micro-expressions from the video stream;
\item \textit{text-based emotion recognition}, which performs semantic sentiment analysis on the transcribed spoken dialogue.
\end{enumerate*}

When enabled, \textit{emotion integration} follows a late-fusion scheme: each channel provides an emotion prediction, reliability weights are applied across channels, and one aggregated emotion label is produced. This single fused label is then forwarded to the LLM context, rather than passing three potentially conflicting emotion outputs independently.
Let $m \in \{v,f,t\}$ denote the voice, facial, and text channels, and let $p_m(e)$ be the confidence assigned by channel $m$ to emotion class $e$. With non-negative reliability weights $w_m$, the fused score is computed as:
\begin{equation}
S(e)=\frac{\sum_{m \in \{v,f,t\}} w_m\,p_m(e)}{\sum_{m \in \{v,f,t\}} w_m},
\label{eq:emotion_fusion}
\end{equation}
and the final emotion sent to the reasoning module is selected by
\begin{equation}
\hat{e}=\arg\max_e\, S(e).
\label{eq:emotion_argmax}
\end{equation}

\paragraph{Pose Monitoring}
Task monitoring verifies, in real time, that the user is correctly performing each stretching exercise and holding the required pose for the prescribed duration. It relies on \textit{MediaPipe Pose}, a real-time skeletal estimation system that detects body landmarks from a single RGB frame, such as the shoulders, elbows, wrists, hips, and ankles. In the current prototype, pose completion is determined through rule-based geometric criteria defined separately for each exercise:

\begin{itemize}[itemsep=5pt]
\item \textit{Arms above the head.} The system checks that both wrists and both elbows have a $y$-coordinate strictly lower than the nose landmark ($y_\text{wrist} < y_\text{nose}$ and $y_\text{elbow} < y_\text{nose}$, where a lower $y$ numerical value corresponds to a position physically higher up because of the top-down image coordinate system), and that the horizontal distance between the two wrists remains below a threshold $w_\text{max}$, ensuring that the arms are raised and approximately centred above the head.

\item \textit{Touch-your-toes.} The system computes the Euclidean distance between each wrist landmark and each ankle landmark:
\begin{equation}
d_{ij} = \sqrt{(x_{\text{wrist},i} - x_{\text{ankle},j})^2 + (y_{\text{wrist},i} - y_{\text{ankle},j})^2}\,,
\label{eq:toe_distance}
\end{equation}
and accepts the pose when at least one wrist--ankle pair is sufficiently close.

\item \textit{Lateral trunk lean (left / right).} The system computes the midpoint of the shoulders $M_s$ and the midpoint of the hips $M_h$, then calculates the signed trunk inclination angle:
\begin{equation}
\alpha = \arctan\!\left(\frac{M_{s,x} - M_{h,x}}{M_{h,y} - M_{s,y}}\right),
\label{eq:trunk_angle}
\end{equation}
where the $y$-axis inversion accounts for OpenCV's top-down coordinate origin. The sign and magnitude of $\alpha$ indicate whether the user is leaning left or right.
\end{itemize}

At each valid frame, the hold timer is incremented by the frame period $\Delta t$, capped at the target duration $T$. If the user remains in an invalid pose for longer than a reset tolerance $t_{\text{reset}}$, the timer is reset and corrective spoken feedback is issued via the text-to-speech channel. Once the timer reaches $T$, the task monitor emits a success event that propagates toward the Action Execution and Planning and Adaptation blocks in Figure~\ref{fig:overall_pipeline}.

For the lateral lean exercise, two independent timers track the left-side and right-side hold durations; both must reach the target duration before the exercise is marked complete.

All pose checks are geometric and frame-specific, requiring no additional training data beyond the pre-trained pose estimation model.

\paragraph{Speech-to-Text}
The speech-to-text component converts spoken user utterances into textual input that can be integrated with visual and task-state information. Its role is to provide a language interface for intent interpretation, clarification requests, and adaptive dialogue decisions during the stretching routine. In practice, the module takes speech audio as input and outputs a text transcription that is merged into the context package consumed by the reasoning module.

Perception and interaction outputs are structured into a unified context package that can include detected objects, task execution status, dialogue input, and optional affect cues. This package is produced by the Perception block and transmitted to the Reasoning block in Figure~\ref{fig:overall_pipeline}.

\subsection{Reasoning Module}

The \textit{Reasoning} block receives the context package from the \textit{Perception} block and sends actionable output to the \textit{Action Execution} block. The reasoning module serves as the central decision-making component, responsible for determining the robot's next action by combining perception outputs, prior knowledge, and the predefined stretching routine. Its operation follows a hybrid approach: the knowledge graph provides structured contextual grounding, while the LLM supports high-level reasoning, contextual adaptation, and natural-language interaction.

\paragraph{Context Integration and Knowledge Graph Retrieval}

Each time the user speaks, the reasoning module compiles an updated context package including:
\begin{enumerate*}
\item the list of objects detected by perception,
\item the aggregated emotional state (single fused label when multimodal affect is enabled),
\item the current exercise progress status,
\item the latest speech-to-text transcription.
\end{enumerate*}

\begin{table*}[!th]
\centering
\small
\caption{Representative examples of entities and relations in the internal knowledge graph.}
\label{tab:kg_examples}
\small
\begin{tabular}{p{2.2cm} p{2.2cm} p{10.5cm}}
\toprule
\textbf{Entity} & \textbf{Type} & \textbf{Example relations} \\
\midrule
\texttt{Banana} & \texttt{Object} &
\texttt{affords} $\rightarrow$ \texttt{EatBanana}; \\
& & \texttt{used\_for} $\rightarrow$ \texttt{QuickEnergyBoost}; \\
& & \texttt{is\_relevant\_when} $\rightarrow$ \texttt{Fatigue}; \\
& & \texttt{is\_a} $\rightarrow$ \texttt{Food} \\

\texttt{DrySweat} & \texttt{Action} &
\texttt{requires} $\rightarrow$ \texttt{Towel}; \\
& & \texttt{helps\_with} $\rightarrow$ \texttt{Comfort} \\

\texttt{Sweating} & \texttt{PhysicalState} &
\texttt{indicates} $\rightarrow$ \texttt{Exertion}; \\
& & \texttt{motivates} $\rightarrow$ \texttt{DrySweat} \\

\texttt{ExerciseSession} & \texttt{Routine} &
\texttt{contains} $\rightarrow$ [\texttt{ArmRaise}, \texttt{ToeTouch}, \texttt{LeanLeftRight}]; \\
& & \texttt{goal} $\rightarrow$ \texttt{BodyRelaxation} \\

\texttt{ToeTouch} & \texttt{Stretch} &
\texttt{targets} $\rightarrow$ \texttt{Hamstrings}; \\
& & \texttt{requires\_flexibility} $\rightarrow$ \texttt{Moderate}; \\
& & \texttt{can\_cause} $\rightarrow$ \texttt{LowerBackStrain} \\

\texttt{Pain} & \texttt{Discomfort} &
\texttt{suggests} $\rightarrow$ \texttt{StopExercise}; \\
& & \texttt{can\_be\_detected\_by} $\rightarrow$ \texttt{TouchingAffectedArea} \\
\bottomrule
\end{tabular}
\end{table*}

To provide the LLM with meaningful background knowledge, this raw context is enriched with semantic relations retrieved from a KG. A KG is a structured representation of entities and their relations, used here to encode contextual and actionable knowledge relevant to assistive robot behavior \cite{Hogan2021,Speer2017}. Unlike unstructured text, a KG enables a system to reason over entities and their connections in a transparent, interpretable way.

The internal KG was manually constructed for the assistive stretching domain and contains entities describing user-related states, environmental objects, actions, exercises, and higher-level contextual knowledge. For example, \texttt{Banana} is represented as an \texttt{Object} linked to \texttt{EatBanana}, \texttt{QuickEnergyBoost}, and \texttt{Fatigue}, while \texttt{ExerciseSession} is represented as a \texttt{Routine} containing the stretches \texttt{ArmRaise}, \texttt{ToeTouch}, and \texttt{LeanLeftRight}. Likewise, \texttt{Pain} is represented as a \texttt{Discomfort} concept linked to the action \texttt{StopExercise}.

Table~\ref{tab:kg_examples} provides representative examples of entities and relations from the current KG.

The current KG therefore mixes several kinds of assistive knowledge: object affordances, user-state cues, action recommendations, exercise-specific information, and simple routine structure. Rather than relying on a single strict ontology layer, the prototype uses a task-oriented representation designed to support retrieval during runtime prompting.

During reasoning, the system extracts context-relevant entities from the current interaction, including detected objects and user-state descriptions. It first checks whether these entities are covered by the internal KG. If an entity is found, its associated relations are retrieved and added to the LLM context. If not, the system falls back to ConceptNet and retrieves a small subset of task-relevant commonsense relations. Retrieved relations are then serialized into the LLM context and used to ground adaptation decisions in explicit symbolic links, such as state-to-action relations (e.g., \texttt{Sweating} $\rightarrow$ \texttt{DrySweat}) or object-to-action relations (e.g., \texttt{Banana} $\rightarrow$ \texttt{EatBanana}).

\paragraph{LLM-Based Decision-Making}

At each interaction cycle, the reasoning module integrates the current perceptual context with knowledge from the internal KG and, when needed, from ConceptNet as a fallback source for entities not covered by the curated graph. The retrieved information is then filtered to retain only task-relevant relations before being passed to the primary LLM. The model analyzes the resulting context to produce a single context-sensitive decision, expressed in a constrained action format that the downstream execution layer can interpret.

For example, if the user reports fatigue while a water bottle is detected in the environment, the retrieved knowledge may connect \texttt{Fatigue} with \texttt{DrinkWater}. This relation is then included in the reasoning context, allowing the LLM to generate an action such as \texttt{POINT\_WATER} accompanied by a supportive utterance. In this way, structured knowledge helps connect perceived user state, environmental affordances, and robot action selection.

The LLM adheres to predefined action prefixes in its outputs:
\begin{itemize}
\item \texttt{NEXT\_EXERCISE:} instructs the system to move to the following stretch.
\item \texttt{POINT\_<OBJECT>:} directs the robot to point to a detected object.
\item \texttt{STOP\_ROUTINE:} ends the stretching session.
\end{itemize}

A secondary verification stage checks response format, tone, and contextual coherence before execution. In the reported pilot sessions, verifier-approved output was emitted as adaptive instructions and feedback to the Action Execution block in Figure~\ref{fig:overall_pipeline}.

\subsection{Action Execution Module}

The \textit{Action Execution} 
% block in Figure~\ref{fig:overall_pipeline}. The action execution 
module grounds high-level reasoning decisions in concrete robot behavior. It acts as the interface between symbolic instructions and the physical world, ensuring that every action is carried out in a synchronized, safe, and comprehensible manner.

\paragraph{Robot Arm Control}

% \paragraph{Command Interpretation}
Each output from the reasoning stage has a dual format: a structured action prefix and a natural-language sentence. The action prefix encodes the intended behavior in machine-readable form, while the text provides a human-friendly explanation, enabling command interpretation.

Once a symbolic command such as \texttt{POINT\_<OBJECT>} is issued, the system translates it into a Cartesian target pose for the robot. A feasible joint-space trajectory is then computed to reach this target while avoiding collisions. The resulting trajectory is smoothed to ensure continuous and stable motion, and is then executed by the robotic arm. Once the target pose is reached within tolerance, an ``execution finished'' event is published back to the reasoning module.

\paragraph{Text-to-Speech}
In parallel with motion control, text-to-speech (TTS) converts the LLM's natural-language output into spoken instructions. The timing of speech and movement is carefully coordinated to enhance clarity and naturalness of the interaction. In socially interactive settings, coordinated turn-taking and expressive feedback can improve the readability and quality of the interaction \cite{Breazeal2003}. The system supports multiple voices, ranging from simple rule-based synthesis to neural voices \cite{VanDenOord2016}, reinforcing engagement during the interaction.

\subsection{Planning \& Adaptation Module}

% In Figure~\ref{fig:overall_pipeline}, this module corresponds to 
The \textit{Planning and Adaptation} module receives control feedback from the \textit{Action Execution} block and maintains the current exercise state for the next cycle. In this way, the planning and adaptation module ensures that StretchBot delivers a structured yet responsive stretching session.

\paragraph{Routine State Tracker}
At its core, the system relies on a predefined exercise script composed of a sequence of stretching primitives (e.g., arm stretch, side bend, neck rotation). Each primitive specifies the expected body posture, a target hold duration, and the transition to the next exercise. This script acts as a baseline plan that guarantees coverage and maintains a clear flow.

\paragraph{Context Updater}
On top of this baseline, the system updates the routine flow based on the user's state and surrounding context. For example:

\begin{itemize}
\item If the user reports fatigue, the system may suggest a brief pause or recommend a coffee break while maintaining the overall sequence.
\item If the user expresses discomfort with a given stretch, the system can substitute a gentler alternative or skip to the next exercise.
\item If relevant objects are detected in the environment, the system proactively offers suggestions (e.g., ``I see a bottle of water nearby—would you like a sip?'').
\end{itemize}

These adaptations occur within the framework of the global routine, ensuring that the session remains coherent and goal-directed while accommodating individual preferences and states. Together, the four blocks in Figure~\ref{fig:overall_pipeline} describe one complete adaptation cycle that readers can follow in parallel with this section.

\section{Implementation}
\label{sec:implementation}

\subsection{Software Stack}

StretchBot is built on a modular software architecture using Robot Operating System 2 (ROS2)~\cite{Macenski2022} to coordinate communication between perception, reasoning, and execution components. Key libraries include:

\begin{itemize}
\item \textbf{Perception:} YOLOv8n (``You Only Look Once'', version 8, nano variant) \cite{YOLOv8} for real-time object detection; MediaPipe Pose \cite{Bazarevsky2020} for skeleton landmark extraction and pose verification; DeepFace \cite{Serengil2021} for facial emotion recognition; HuggingFace Transformers pipelines \cite{Wolf2020} for voice-segment and transcript-level emotion analysis.
\item \textbf{Reasoning:} Custom Python modules for KG retrieval and context integration; OpenRouter-served open-weight models for LLM-based planning; optional verifier integration via the HuggingFace Inference Client~\cite{HuggingFaceInference}.
\item \textbf{Execution:} ROS2 command publishing for symbolic actions; local \textbf{gTTS}~\cite{gTTS} speech synthesis; optional \textbf{Vosk} STT (offline)~\cite{Vosk2021} and optional expressive TTS through a HuggingFace Gradio space~\cite{Abid2019}.
\item \textbf{Spatial Grounding:} In the pilot, object positions were fixed and predefined. The repository architecture supports optional ArUco marker-based localization (4×4 50-dictionary, OpenCV~\cite{Bradski2000}) for future metric 3-D pose estimation when dynamic object tracking is required.

\end{itemize}

A key architectural decision is the \textbf{concurrent execution} of LLM reasoning and pose monitoring. The LLM interaction runs in a dedicated Python thread, while the main thread executes the real-time MediaPipe pose-tracking loop at 30\,Hz. An \texttt{exercise\_done} flag and a \texttt{stop\_flag} dictionary coordinate termination between the two threads. In the current pilot script, interaction input is text-based; STT can be enabled as a runtime variant.

\subsection{Parameter Values}

The current pilot prototype uses a small set of manually chosen parameter values for pose monitoring and exercise progression. These values were empirically selected to provide a simple and stable proof-of-concept behavior in the controlled evaluation setting.

For the \textit{arms above the head} exercise, the horizontal wrist-distance threshold was set to $w_{\text{max}} = 0.3$ in normalized image coordinates. For \textit{touch-your-toes}, the wrist--ankle proximity threshold was set to $d_{ij} < 0.4$. For the \textit{lateral trunk lean} exercise, the user was classified as leaning right when $\alpha > 15^\circ$ and leaning left when $\alpha < -15^\circ$.

The pose-monitoring loop operated at 30\,Hz, corresponding to a frame period of $\Delta t = 1/30\,\text{s}$. The target hold duration for each pose was set to $T = 5\,\text{s}$, and the reset tolerance for sustained invalid posture was set to $t_{\text{reset}} = 40\,\text{s}$. These values were intended to provide readable feedback and robust completion detection in the small-scale pilot, rather than to optimize performance across users.

\subsection{Reasoning Module Details}

The primary LLM used in the pilot implementation was \texttt{deepseek-r1-0528-qwen3-8b}, accessed through OpenRouter~\cite{Guo2025}. It is instructed through a structured system prompt that provides context on the user, the stretching routine, detected objects, and the user's emotional state. The prompt explicitly instructs the model to use predefined action prefixes, namely \texttt{NEXT\_EXERCISE}, \texttt{POINT\_<OBJECT>}, and \texttt{STOP\_ROUTINE}, and to adapt its tone based on the user's emotional state. This format ensures consistent machine-readable output while maintaining natural language quality. A full prompt example is provided in Appendix~\ref{app:prompt_example}.

% \subsection{Knowledge Representation}

The internal knowledge graph is implemented as a lightweight JSON-based resource parsed at runtime into Python dictionaries. Each entry represents an entity and stores (i) a \texttt{type} field and (ii) a \texttt{relations} dictionary that maps relation names to one or more linked concepts. This representation was chosen as a simple alternative to a full RDF triple store while remaining sufficient for the current prototype.

\subsection{Hardware Setup}

The prototype was developed for indoor assistive interaction with a camera stream, speech output, and ROS2-based action dispatch to robotic control software. In the reported pilot, object positions were predefined and fixed. The repository architecture supports optional ArUco marker-based localization~\cite{GarridoJurado2014}, where square fiducial markers placed in the environment are detected by the camera and used to estimate object or scene pose for spatial grounding.

% \subsubsection{Cost and Compute Footprint}

The pilot prototype was intentionally developed under low-budget constraints. In practice, the robotic arm accounted for the majority of the hardware cost. At the same time, most of the remaining software stack relied on free or open-source components (ROS2, OpenCV, MediaPipe, and open-weight or API-accessed language models). The runtime also operated on a modest, non-high-end PC, indicating that the architecture is feasible without specialized compute hardware for proof-of-concept deployment.

\section{Pilot Evaluation}
\label{sec:evaluation}

\subsection{Evaluation Design}

We conducted an exploratory user study that compared a scripted baseline with the adaptive system. The study design was a pilot comparison where participants experienced both systems. This evaluation is intended as an initial feasibility and user-feedback study, not as a confirmatory validation of performance differences.

\subsection{Participants and Protocol}

Three adult participants took part in the pilot study. Each participant completed two stretching sessions, each lasting 8--10 minutes, under two conditions presented in randomized order: (1) a \textbf{scripted} baseline, in which the robot followed a fixed exercise sequence without perception-based adaptation, and (2) an \textbf{adaptive} condition, in which the robot adjusted the guidance based on user state and environmental context.

A two-minute break separated the two conditions. Participants were informed of the study's purpose and signed informed consent forms. The study was conducted in a controlled laboratory environment with standard objects (a coffee mug, a water bottle, a banana, and a chair) present.

\subsection{Quantitative Metrics}

After each condition, participants completed a custom questionnaire rating the robot's performance on six dimensions:

\begin{enumerate}
\item \textbf{Clarity:} How clear were the robot's instructions? (1–5 Likert scale)
\item \textbf{Comfort:} How comfortable were you during the routine? (1–5)
\item \textbf{Adaptability:} How well did the robot adapt to your state and needs? (1–5)
\item \textbf{Trust:} How much did you trust the robot's decisions? (1–5)
\item \textbf{Naturalness:} How natural and human-like was the robot's behavior? (1–5)
\item \textbf{Object Relevance:} How relevant were the robot's object suggestions? (1–5, N/A for scripted condition)
\end{enumerate}

\subsection{Qualitative Feedback}

After completing both conditions, participants participated in semi-structured interviews (15–20 minutes) discussing:
\begin{enumerate*}    
\item Which version they preferred and why;
\item Perceived strengths and weaknesses of each approach;
\item Suggestions for improvement;
\item Overall sense of engagement and repeated-use intentions.
\end{enumerate*}

Qualitative feedback from participant interviews was summarized and used to identify recurring perceptions of strengths, limitations, and user preferences across scripted and adaptive interaction modes.

\begin{figure*}[t]
\centering
\includegraphics[width=0.8\textwidth]{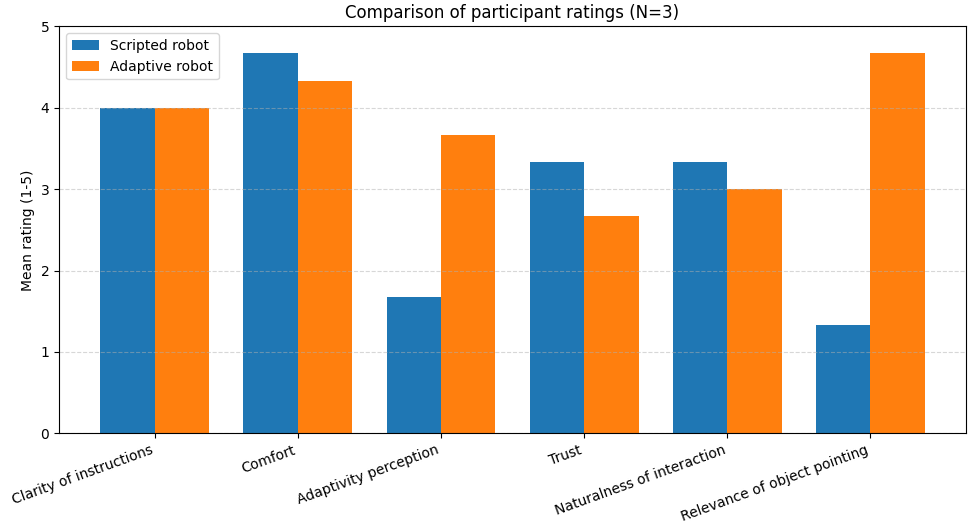}
\caption{Comparative results between scripted and adaptive conditions across the main user-evaluation dimensions.}
\label{fig:comparative_results}
\end{figure*}

\section{Results}

\label{sec:results}

\subsection{Quantitative Results}

Because the study involved only three participants in a controlled proof-of-concept setting, the results should be viewed as preliminary evidence rather than definitive confirmation. They provide useful early indications of user-perceived trends and trade-offs, but require validation in larger, more diverse evaluations.

Figure~\ref{fig:comparative_results} provides a visual comparison of mean participant ratings in the scripted and adaptive conditions. In the following, $M$ denotes the mean rating and $SD$ denotes the standard deviation.

\textbf{Clarity:} Both conditions received the same mean rating for clarity (Scripted: $M = 4.00, SD = 0.00$; Adaptive: $M = 4.00, SD = 1.00$), suggesting that both interaction modes were generally perceived as clear in this pilot.

\textbf{Comfort:} Comfort ratings were slightly higher for the scripted condition (Scripted: $M = 4.67, SD = 0.58$; Adaptive: $M = 4.33, SD = 0.58$), indicating that the scripted version may have been perceived as somewhat easier or more relaxing to follow.

\textbf{Adaptability:} The adaptive condition received notably higher ratings for perceived adaptability (Adaptive: $M = 3.67, SD = 1.15$; Scripted: $M = 1.67, SD = 1.15$). Although this difference is based on a very small sample, it is consistent with the adaptive system's intended role.

\textbf{Trust:} Trust ratings were somewhat higher for the scripted condition (Scripted: $M = 3.33, SD = 0.58$; Adaptive: $M = 2.67, SD = 1.15$), suggesting that greater predictability may have supported user confidence in this pilot setting.

\textbf{Naturalness:} Naturalness ratings were also slightly higher for the scripted condition (Scripted: $M = 3.33, SD = 0.58$; Adaptive: $M = 3.00, SD = 0.00$), indicating that adaptive behavior did not automatically translate into a more natural interaction experience.

\textbf{Object-pointing relevance:} Ratings for object-pointing relevance were markedly higher in the adaptive condition (Adaptive: $M = 4.67, SD = 0.58$; Scripted: $M = 1.33, SD = 0.58$), suggesting that context-aware object references were perceived as substantially more relevant when generated by the adaptive system.

Overall, the quantitative findings suggest a clear trade-off between adaptability and interaction smoothness. In this pilot, the adaptive condition was rated more highly for perceived adaptability and object-pointing relevance, while the scripted condition was rated slightly more favorably for comfort, trust, and naturalness. Clarity remained comparable across both conditions. These results suggest that adaptive guidance can improve contextual relevance, but may also introduce interaction costs that affect how natural or comfortable the experience feels.

\subsection{Qualitative Findings}

The qualitative analysis aimed to capture participants' experiences of the scripted and adaptive conditions and to identify the main perceived trade-offs between them. Although the sample was small, the interviews provide useful insight into how users interpreted adaptation, smoothness, and cognitive load during the interaction.

Two participants preferred the scripted condition, mainly because it felt smoother, more predictable, and easier to follow. They described it as more relaxing and requiring less mental effort, since they could simply follow the routine without having to respond to additional prompts or make decisions during the session. By contrast, one participant preferred the adaptive condition, describing it as more personalized and more responsive to contextual cues. In particular, context-specific recommendations and object-aware prompts gave the impression that the robot was more attentive to the user's state and needs.

These interviews also highlight an important trade-off. The adaptive condition was perceived as more thoughtful and context-aware, but this additional responsiveness sometimes came at the cost of interaction flow. For some users, adaptive prompts introduced extra cognitive load and occasionally made the session feel less natural or less relaxing. The scripted condition, while less personalized, benefited from greater continuity and predictability. Across both conditions, participants reported feeling safe, and no participant reported unsafe suggestions during the pilot sessions.

\subsection{Technical Observations}

The following observations emerged from the pilot implementation. However, given the small sample size ($N=3$) and a single experimental session per participant, these findings should be viewed as exploratory rather than confirmatory:

% \begin{itemize}
% \item 
\noindent \textbf{Comparison Between Internal KG and ConceptNet:} Internal inspection of pilot runs suggested that many responses were grounded in internal KG, while a subset required ConceptNet fallback. This indicates that the curated KG covered common stretching-domain concepts, with broader commonsense retrieval helping for less common inputs.

% \item 
\noindent \textbf{LLM Verification Effects:} To better characterize the impact of the verifier, we tracked how often first-pass LLM outputs were modified before execution. Most edits were lightweight corrections (action-prefix normalization, formatting repair, and tone adjustment), while full semantic rewrites were less frequent. This second-pass filtering improved command consistency and conversational appropriateness, but introduced additional latency because each response required an extra verification step.

% \item 
\noindent \textbf{Response Latency Under Low-Budget Setup:} In the pilot's low-budget configuration (modest local machine and API-based reasoning), response delays were sometimes noticeable. They could occasionally reach roughly 3--10 seconds before coach feedback was produced. Participants generally tolerated this in a proof-of-concept context, but such pauses can reduce immersion during guided interaction.

% \end{itemize}
\section{Discussion}
\label{sec:discussion}

This discussion interprets the pilot findings as exploratory signals rather than full empirical validation. Their main value is to show that the prototype can be deployed in interactive sessions and that participants noticed meaningful differences between scripted and adaptive behavior; no adverse events or participant-reported safety incidents were observed during the pilot sessions.

\subsection{Key Findings}

\textbf{Dimension-Specific Effectiveness and Verification Trade-off:} In the adaptive condition, the system showed promise for adaptability and environmental grounding, suggesting that LLM-KG integration can support context-aware decision-making. % in this pilot setting. 
However, this trend came with interaction costs for some users: two of three participants reported that the scripted condition felt more relaxing and natural, suggesting higher cognitive load in adaptive interactions. A likely contributor is the additional verification stage, which improved response consistency and conversational appropriateness but added latency and occasionally reduced flow smoothness. These results remain exploratory and require validation in larger-scale studies.

\noindent \textbf{Actionable Knowledge Representation:} In our pilot study, the internal KG appeared effective for common scenarios (fatigue, hydration, object affordances) but still required commonsense fallback for some novel situations. This supports the hypothesis that robust adaptation may require a hybrid representational backbone combining curated symbolic knowledge with broader commonsense resources, though larger studies are needed to assess generalizability.

\subsection{Discussion of Design Trade-offs and Practical Implications}

A central design choice in StretchBot is the \textbf{balance between a global scripted baseline and local adaptive branching points}. This hybrid approach offers a middle ground: the global routine ensures exercise coverage and predictability, while local adaptations provide personalization without making the interaction fully reactive or unpredictable. In our pilot, this trade-off was reflected in user feedback. The adaptive condition improved perceived adaptability and object-pointing relevance, whereas the scripted condition was more often described as smoother, easier to follow, and more relaxing.

This pattern suggests that the optimal balance may be user-dependent. Users who prefer predictability may benefit from a script-first design in which adaptation is introduced selectively, while users who value personalization may prefer a more adaptive interaction style with a scripted fallback. In this sense, adaptive assistive guidance should not be treated as uniformly preferable to scripted guidance, but rather as a design space in which different balances between structure and flexibility may suit different users and contexts.

Multimodal perception contributes to this balance, but its role should be interpreted carefully. Combining voice, facial, and text-based emotion signals is intended to reduce susceptibility to single-modality failures by producing one fused affect estimate for the reasoning stage. However, in this pilot, multimodal affect remained an optional context component and was not the dominant adaptation driver in all runs. Emotion recognition also remains an imperfect proxy for actual user state: a user may sound tired while remaining engaged, and facial expressions may be ambiguous or context-dependent. These considerations suggest that affective signals are useful as supplementary cues, but should not be relied upon in isolation for assistive decision-making.

A second important trade-off concerns \textbf{latency}. In the low-budget pilot setup, adaptive responses were often slower than scripted execution, with occasional delays in the 3--10 second range before feedback was produced. For a proof-of-concept stretching routine, this remained manageable, but such pauses can disrupt the flow of interaction and reduce perceived naturalness. A likely contributor is the additional verification stage, which improved response consistency and conversational appropriateness but also added overhead. This highlights a broader point: better reasoning quality does not automatically translate into a better user experience if responsiveness is degraded.

More broadly, the pilot suggests that \textbf{coupling LLMs with structured knowledge representations} may offer a practical path toward more grounded and interpretable planning in assistive robotics. The internal KG appeared effective for common scenarios such as fatigue, hydration, and object affordances, while ConceptNet fallback helped cover less common cases. This supports the view that robust adaptive guidance may benefit from a hybrid representational backbone that combines curated symbolic knowledge with broader commonsense resources. At the same time, the integration remains non-trivial: prompt engineering, knowledge graph design, latency management, and verification policies all affect the final interaction quality. For assistive robotics specifically, a key implication is that personalization and safety are not orthogonal. By grounding LLM reasoning in domain knowledge and user perception, and by constraining execution through explicit action formats and verification, the system aims to support adaptation while preserving conservative interaction boundaries.

\subsection{Future Directions and Deployment Recommendations}

Based on this pilot, we identify several directions for both practical deployment and future research.

First, adaptive assistive systems should \textbf{default to low cognitive load}. A smooth, scripted flow appears valuable for user comfort, and adaptive branching should ideally be introduced only when user state or situational context indicates a clear benefit. Relatedly, future systems should expose adaptation controls more explicitly, for example by allowing users to choose between a more scripted and a more adaptive interaction mode, or to switch modes during a session.

Second, \textbf{safety-critical constraints should remain deterministic} even when higher-level interaction is adaptive. In practice, this means separating hard safety rules (e.g., motion bounds, forbidden actions, emergency interruption conditions) from conversational flexibility. Structured logging of perception signals, KG evidence, and verifier edits would further improve auditability and support post-hoc failure analysis.

Third, \textbf{larger and more diverse evaluations} are needed. Future studies should include more participants, richer user profiles, and longer interaction horizons in order to assess robustness, user adaptation over time, and long-term acceptance. This also opens the way to more nuanced state representations, for example by incorporating biometric or physiological signals alongside emotions and dialogue context.

Fourth, several \textbf{technical extensions} could strengthen the framework. These include improving latency through lighter models, local deployment, or more efficient verification strategies; expanding the internal KG or partially automating its construction; extending the framework to other assistive domains such as rehabilitation or elderly support; and learning user-specific preferences over time. Finally, more formal verification mechanisms could be explored to ensure that adaptive reasoning remains aligned with safety and task constraints even in more open-ended real-world settings.

\section{Conclusion}
\label{sec:conclusion}

This paper presented StretchBot, a neuro-symbolic adaptive robotic coach that integrates multimodal perception, knowledge graph reasoning, and large language models to guide personalized stretching routines. The primary contribution is the system design and implementation of a practical neuro-symbolic pipeline for grounded assistive guidance, while the pilot user study serves as a complementary exploratory assessment of feasibility and perceived interaction trade-offs.

Through a pilot user study with three participants, we explored the feasibility and user experience of adaptive versus scripted robot guidance in a stretching domain. Preliminary findings suggest that the adaptive system shows promise for improving perceived contextual relevance and environmental awareness, while scripted interaction remains competitive for perceived smoothness and predictability. These exploratory results motivate further research into user-centered design of adaptive assistive robots and suggest that the optimal interaction paradigm may be user-dependent.

\subsection{Broader Impact}

Adaptive assistive robots have significant potential to improve health outcomes and quality of life for aging populations and individuals with mobility or cognitive limitations. However, responsible deployment of such systems requires careful attention to user autonomy, informed consent, transparency in decision-making, and mechanisms for user control and override. This work represents a step toward principled, user-centered design of adaptive assistive robots, though real-world deployment requires ongoing engagement with ethics boards, healthcare practitioners, and end-users to ensure alignment with societal values and individual needs.

\backmatter

\begin{appendices}

\section{Example System Prompt}
\label{app:prompt_example}

\begin{quote}
\small\ttfamily
You are StretchBot, a friendly and empathetic robot coach guiding a human through a safe and supportive morning stretching routine and normal conversations.\\[0.3em]

Your stretching plan is:\\
Stretch your arms above your head for 5 seconds\\
Touch your toes for 5 seconds\\
lean left and right for 5 seconds each\\[0.3em]

Current exercise: \{current\_exercise\}\\
Next exercise: \{next\_exercise\}\\[0.3em]

Context:\\
\{context\_description\}\\
\{history\_str\}\\[0.3em]

Relevant commonsense knowledge:\\
\{formatted\_kg\}\\[0.3em]

Your instructions:\\
1. Analyze the user's current state based on the context and recent dialogue.\\
2. If the user feels fine, gently propose moving on to the next exercise: \{next\_exercise\}, with clear instructions.\\
3. If the user seems tired, tense, or has previously expressed discomfort, suggest help (e.g., break, water, encouragement), but do NOT repeat offers they already refused.\\
4. Always prioritize the user's most recent response.\\
5. Do not offer the same help more than once unless the user expresses a new need.\\
6. Speak with short, warm, and simple sentences. Use friendly language. Congratulate or encourage when appropriate.\\
7. Ask a caring question if you think the user may be struggling or needs support.\\
8. If the user is asking a question, always answer it with something related to that question, even if it is not related to the current exercise.\\
9. If you want the robot to point to an object detected in front of it (for example, a glass, a banana, or a towel), start your Output line with: POINT\_<OBJECT> (for example: POINT\_GLASS, POINT\_BANANA, POINT\_TOWEL), then continue your sentence naturally.\\
10. If the user wants to stop the stretching routine, or if you think it is necessary to stop for safety or well-being, or if the routine is over, start your Output line with: STOP\_ROUTINE --- this will be your final message to the user.\\[0.3em]

- You will receive a line like `Exercise status: success' or `Exercise status: not yet' in the context.\\
- If the status is `success', congratulate the user and propose moving to the next exercise only if the user succeeded.\\\\
- If the status is `not yet', encourage the user to keep trying and give advice.\\[0.3em]

IMPORTANT:\\
- If it's appropriate to start the next exercise, begin your Output line with: NEXT\_EXERCISE:\\
- If you want the robot to point to an object, begin your Output line with: POINT\_<OBJECT> (replace <OBJECT> by the object name in English and uppercase, e.g., POINT\_GLASS).\\
- If the user wants to stop or you decide to stop the stretching routine because you have no more exercises left, begin your Output line with: STOP\_ROUTINE, this will be your final message to the user.\\
- Otherwise, respond naturally and empathetically to the user.\\[0.3em]

Format your reply like this and only one time:\\[0.3em]

Reasoning:\\
<step-by-step reasoning>\\[0.3em]

Output: <what the robot should say or ask next in 1-2 sentences>
\end{quote}

\end{appendices}

\bmhead{Acknowledgements}

We thank all study participants for their time and feedback. We acknowledge support from the Vrije Universiteit Amsterdam and the Master's program in IRobotics at the University of Montpellier.
We thank Shujian Yu for helpful discussions and feedback.

\section*{Statements and Declarations}

\subsubsection*{Funding} No external funding was received for this research beyond standard university support.

\subsubsection*{Competing Interests} The authors have no relevant financial or non-financial interests to disclose.

\subsubsection*{Ethics Approval and Consent to Participate} This pilot involved adult volunteers in a non-clinical stretching-assistance scenario. Participants were informed about study procedures and provided consent prior to participation.

\subsubsection*{Consent for Publication} Participants consented to the use of anonymized study results in research dissemination. Personal and directly identifying information was removed from reported data.

\subsubsection*{Data and Code Availability}
Code and documentation for the prototype are available at \url{https://github.com/Lucavogel/adaptive_robot_planning}.

Anonymized pilot-level quantitative summaries and de-identified qualitative excerpts are available from the corresponding authors upon reasonable request via email, subject to privacy constraints.

Raw audio/video recordings are not shared due to participant privacy.

\subsubsection*{Materials Availability} Not applicable.

\bibliography{references/paper_stretchbot}

\end{document}